\documentclass{article}

\usepackage{arxiv}

\usepackage[utf8]{inputenc} 
\usepackage[T1]{fontenc}    
\usepackage{hyperref}       
\usepackage{url}            
\usepackage{booktabs}       
\usepackage{amsfonts}       
\usepackage{nicefrac}       
\usepackage{microtype}      
\usepackage{lipsum}		
\usepackage{graphicx}
\usepackage{doi}
\usepackage{multirow}

\title{Agave crop segmentation and maturity classification with deep learning data-centric strategies using very high-resolution satellite imagery}

\author{ \hspace{1mm}
Abraham Sánchez, Raúl Nanclares, \href{https://orcid.org/0000-0001-9969-1339}{ \includegraphics[scale=0.06]{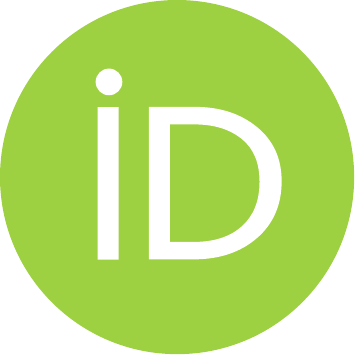}\hspace{1mm} Alexander Quevedo} \\
	Coordinación General de Innovación Gubernamental,
	del Gobierno de Jalisco\\
	Guadalajara Jalisco, México\\
	\texttt{abraham.sanchez, raul.nanclares, alexander.quevedo@jalisco.gob.mx} \\
	\And
    Ulises Pelagio \\
	Centro de Investigaciones en Geografía Ambiental \\
	Universidad Nacional Autónoma de México \\
	Morelia, Michoacán, México\\
	\texttt{ujimenez@pmip.unam.mx} \\
	\And
    Alejandra Aguilar \\
	Secretaría de Medio Ambiente y Desarrollo Territorial \\
	Gobierno de Jalisco\\
	Guadalajara Jalisco, México\\
	\texttt{alejandra.aguilarramirez@jalisco.gob.mx} \\
	\And
    Gabriela Calvario \\
	 Department of Electronics, Systems, and Informatics \\
	ITESO—The Jesuit University of Guadalajara, Tlaquepaque\\
	Tlaquepaque, Jalisco, México\\
	\texttt{gabriela.calvario@iteso.mx } \\
	\And
	\href{https://orcid.org/0000-0002-5123-4881}{\includegraphics[scale=0.06]{orcid.pdf}\hspace{1mm} E. Ulises Moya-Sánchez}\thanks{Corresponding Author.} \\
	Coordinación General de Innovación Gubernamental/ Postgrado en ciencias computacionales UAG\\
	Gobierno de Jalisco/ Universidad Autónoma de Guadalajara,
	Guadalajara, Jalisco, México\\
	\texttt{eduardo.moya,@jalisco.gob.mx,@edu.uag.mx} \\
}

\renewcommand{\shorttitle}{}

\hypersetup{
pdftitle={Agave crop segmentation and maturity classification with deep learning data-centric strategies using very high-resolution satellite imagery },
pdfsubject={q-bio.NC, q-bio.QM},
pdfauthor={Ulises Moya},
pdfkeywords={First keyword, Second keyword, More},
}

\begin{document}
\maketitle

\begin{abstract}
The responsible and sustainable agave-tequila production chain is fundamental for  the social, environment and economic development of Mexico's agave regions. It is therefore relevant to develop new tools for large scale automatic agave region monitoring.  In this work, we present a   
\textit{Agave tequilana Weber}  \textit{azul} crop segmentation and maturity classification using very high resolution satellite imagery, which could be useful for this task. To achieve this, we solve real-world deep learning problems in the very specific context of agave crop segmentation such as lack of data, low quality labels, highly imbalanced data, and low model performance. The proposed  strategies go beyond data augmentation and data transfer  combining   active learning and the creation of synthetic images with human supervision.  As a result,  the segmentation performance evaluated with Intersection over Union (IoU) value  increased from 0.72 to 0.90 in the test set.   We also propose a method for classifying agave crop maturity with 95\% accuracy. With the resulting accurate models,  agave production forecasting can be made available for large regions. In addition, some supply-demand problems such excessive  supplies of agave or, deforestation, could be  detected early.
\end{abstract}

\keywords{Large-scale agave segmentation \and Agave maturity classification \and Deep learning}

\section{Introduction}
\label{intro}

The agave-tequila industry is one of Mexico's leading agribusinesses \cite{gallardo2018industria}. Tequila  and  agave production is  particularly important for Jalisco's  socioeconomic development, this state being Mexico's  largest tequila and agave producer~\cite{crt}. 

The development of decision-making support tools for governments and regulators is crucial to facilitate  responsible-sustainable production and minimum support prices for agricultural commodities. For responsible and sustainable  agave production,  the Jalisco government and the Tequila Regulatory Council (TRC) is promoting Agave Responsible Ambient (ARA) ~\cite{ARA} certification. In this context,   agave crop monitoring on a regional (large) scale is imperative to preserve natural biodiversity, reduce deforestation in the region, and decrease social impacts.  

Human supervision of regional (large-size) geographic regions is an exhausting and time-consuming task. Automatic or semi-automatic monitoring techniques  are now available thanks to the next generation  Artificial Intelligence (AI) models, Deep Learning (DL). In recent years, DL techniques have outperformed other techniques in computer vision tasks in the fields of agriculture and environmental monitoring including plant counting~ \cite{kitano2019corn,flores-gonzalez-etal4-2021}, land use/land cover classification~\cite{quevedo-sanchez-nanclares-montoya-martinez-pacho-moya-2022}, and plant disease prediction\cite{bhakta-2022-plant-disease}. 

Despite these impressive results,  DL performance is not guaranteed in all scenarios. For example, it is well known that low-quality images, or low quantities of data or out-of-distribution data can significantly reduce model performance ~\cite{stadelmann2018deepinthewild, dodge2016understanding, moya2021trainable}.  For instance,  we face three main problems in this work: i) little training data,  ii) widely diverse agricultural practices for the different agave crops in the training dataset such as: a lack of data diversity, difficulty in separating  "young" agave crops from other crops with similar plantation patterns, or the  presence of diverse features among the crops (other plants, trees, water reservoirs, buildings, etc.)  and  iii) geographic scale: enormous datasets and large geographic areas with considerable variability in soil types, vegetation, agricultural practices, etc.

In this work, we propose the use of  DL models, data augmentation and AI data-centric techniques \cite{whang2023data-centric} to segment, and classify agave crops into age groups using very high spatial resolution imagery. The entire segmentation tasks  processes  is grouped into three main phases: i) phase 1, using the original training data/labels and data augmentation,  ii) phase 2, improving training data labels, through joint human and model data labeling, and iii) phase 3, adding  synthetic images to the training data sets. In Section \ref{sec:exp}) we provide more details about each phase. 

The main contributions of this work are to: i) generate effective DL models for agave segmentation and agave-maturity classification on a regional scale using satellite imagery; ii) propose a novel strategy combining the (DL) techniques such as transfer learning with human experts to increase training data sets size, and improve data-label quality; iii) compare the performance of some of the most widely used  DL segmentation models: Mask R-CNN \cite{maskrcnn}, Unet++ \cite{unetplusplus}, DeepLabV3+ \cite{deeplabv3plus} and FPN \cite{fpn}. Moreover, compare three  well known lightweight classification models MobileNetV2 \cite{mobilenet}, ResNet18 \cite{resnet} and EfficientNet B0 \cite{efficientnet}. 

According to the results, the proposed data-centric techniques improved segmentation models performance  significantly, raising the Intersection over Union (IoU) score from 0.70 to 0.91 and achieving  95 \% accuracy in agave age-stage classification.

The rest of the paper is organized as follows. In Section \ref{sec:related} the related-work  information is gathered.  The datasets  and the experimental setup are described in
Sections \ref{sec:data} and \ref{sec:exp}. After that, the results and their analysis are presented in Section \ref{sec:results}.  Our conclusions and future work are provided in Section \ref{sec:conclusions}.

\section{Related work}\label{sec:related}

 Most of  previous works related to agave detection or segmentation use UAV (Unmanned Aerial Vehicle) imagery~\cite{flores-gonzalez-etal4-2021,escobar2022unmanned-agave-uav,calvario-alarcon-etal-2020agave, calvario-sierra-multi2017agave}. These UAV images have a ground sampling distance (gsd) of approximately 1-3 cm, higher than WorldView-2's  30-50 cm gsd satellite imagery. Moreover,   these  works have different objectives. For instance,  agave plantation line detection, agave plant detection or plant counting. In contrast, the models in our work,  are designed to evaluate  large geographic regions and segment agave crops. A related work that uses satellite data as input  for agave detection is \cite{garnica2008remotesening-agave}.  In this related work,  the authors affirmed that the agave is mixed with other types of vegetation such as grasslands and their average accuracy performance only achieves $~70\%$. Moreover, the corresponding techniques in  \cite{garnica2008remotesening-agave} are also different, including: Machine Learning (ML), and classic Computer Vision (CV). In Table \ref{tab:previous} we provide more details on each related work. According to the previous information of the related work, it is not possible to  make a direct performance comparison between those works and own.  It is important to remark that our work, to the best of our knowledge, is the only method which combines regional scale data (satellite images) with state-of-the-art DL  and data-centric techniques for agave crop segmentation.

\begin{table}[h]
    \centering
    \begin{tabular}{lccl}
    \toprule
    Work          & Sensor & gsd &  Technique/Task     \\
    \midrule
    Flores et al. \cite{flores-gonzalez-etal4-2021}      & UAV & 0.025 m  & ML,DL/Plant segmentation   \\
    Escobar et al. \cite{escobar2022unmanned-agave-uav} & UAV & 0.026 m  & ML,DL/ Plant detection    \\
    Calvario et al. \cite{calvario-alarcon-etal-2020agave} & UAV & 0.03 m  & CV/Plant counting    \\
    Calvario et al. \cite{calvario-sierra-multi2017agave} & UAV & 0.026 m  & ML/Plant line detection   \\
    Garnica et al. \cite{garnica2008remotesening-agave}      & Landsat 7 &  30 m   &  ML/Crop land detection  \\
    Ours      & WorldView-2 & 0.3-0.5 m   & DL/ Crop segmentation \\
    \bottomrule
    \end{tabular}
    \caption{Details of literature review. Our work is the first to implement DL techniques for crop agave segmentation on a regional scale. }
    \label{tab:previous}
\end{table}

Regarding the proposed data-centric strategy, other works, such as \cite{synthetic}, propose the use of synthetic data to increase the amount of training data. However, the effectiveness of these techniques depends on the nature of the problem and the data distribution of the patterns in the images. In contrast,   we must select different crop shapes, different neighboring constructions, different crop types,  different soil colors, etc. with  visual patterns to similar those of agave crops in order to generate the most representative data.

\section{Data}\label{sec:data}
\begin{figure}[ht]
\begin{center}
\includegraphics[width=0.98\textwidth]{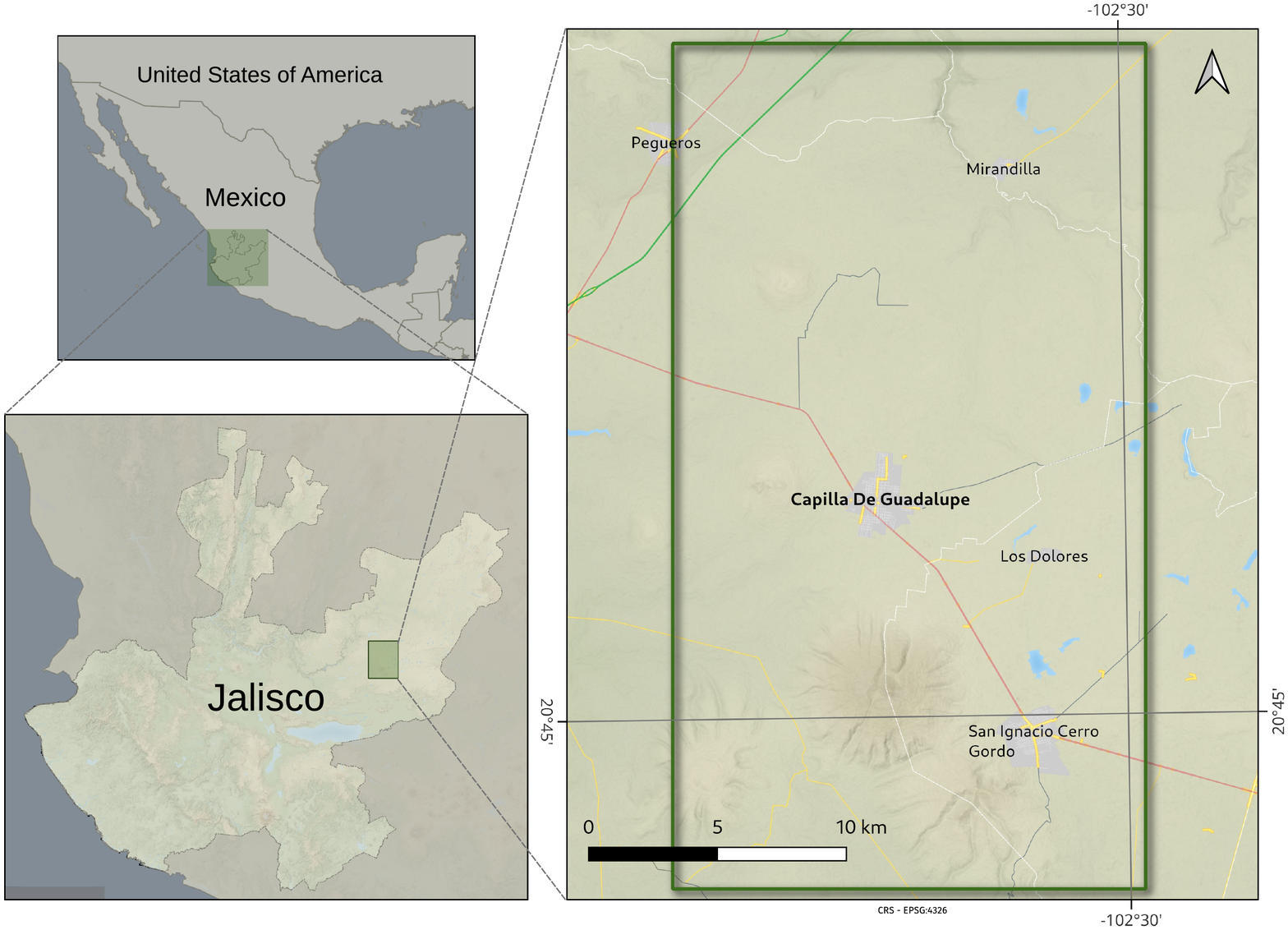}
 \caption{Regional area under study:  Capilla de Guadalupe and San Ignacio Cerro Gordo municipalities of Jalisco, Mexico.}
\label{fig:StudyArea}
\end{center}
\end{figure}

The area under study includes the localities of Capilla de Guadalupe and San Ignacio Cerro Gordo,  in the state of Jalisco, Mexico. Figure \ref{fig:StudyArea}  presents the  area under study in the central west region of Mexico.  The total area of study is  600.7 $km^2$. The  area was divided using a  grid (G)  which resulted in  100 2.5 $\times 2.5$ $ km^2$ Raster (R)-Labels (L) pairs ($P=(R_{G},L_{G})$) associated with the grid  as show in Figure \ref{fig:raster_grid}.

 We used Worldview-2 RGB-images (2019 and 2020)  and their corresponding  labels (digitized by expert photointerpreters using \href{https://www.qgis.org}{QGIS}) as training data. 
 
\subsection{Agave crop segmentation}

\begin{figure}
    \centering
    \includegraphics[width=1.0\textwidth]{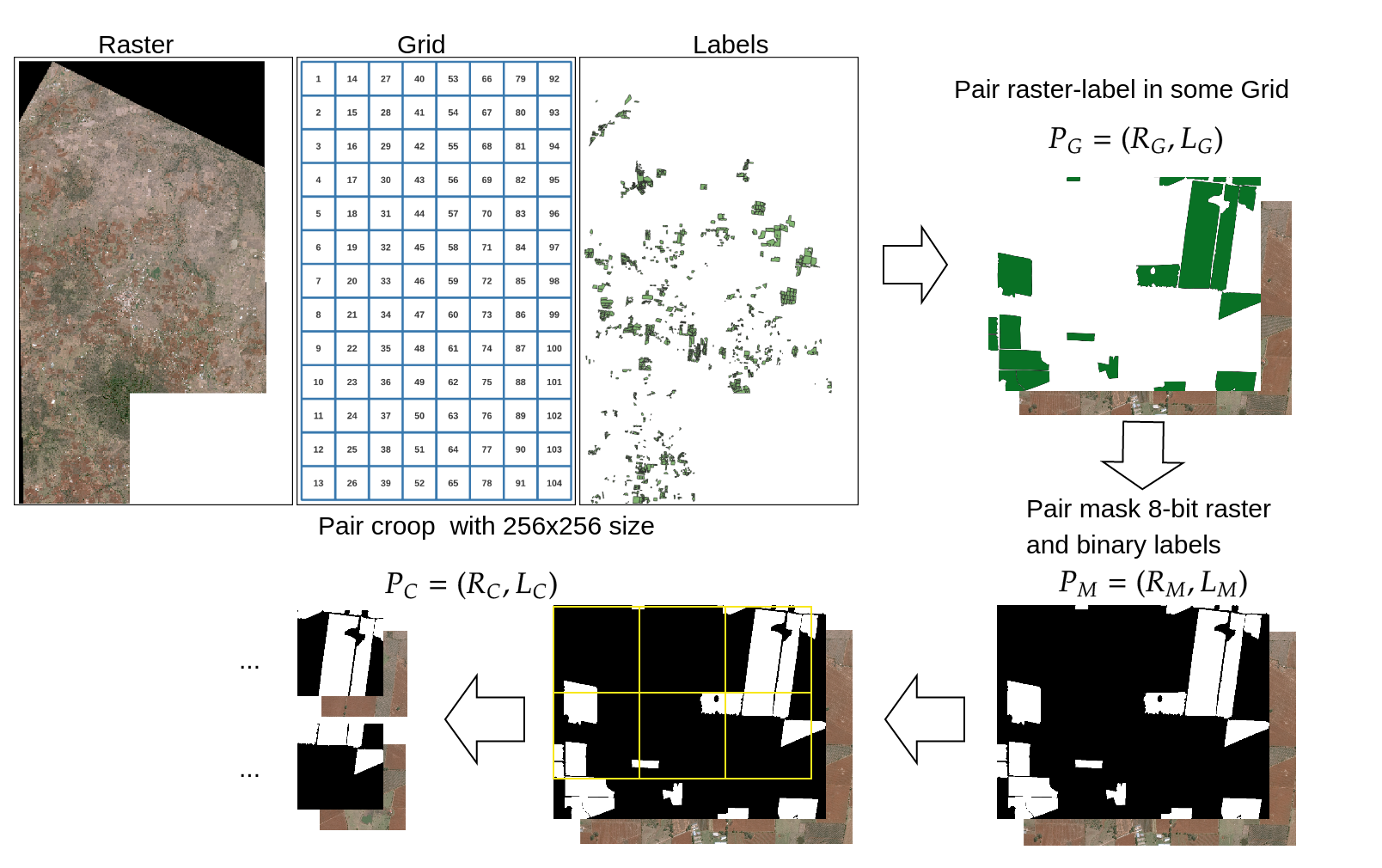}
    \caption{ Agave crop segmentation Pre-processing steps.}
    \label{fig:raster_grid}
\end{figure}

 Worldview-2 data and their labels were grouped in Grid (G) elements generating the pairs $P_{G} = (R_{G},L_{G})$.  Then, a pair mask ($P_{M} = (R_{M},L_{M})$) was generated with an 8-bits re-sampling and a binary raster mask (as segmentation label).  After that,  a 256x256  cropping was applied to  rasters and labels, defined by pair crop $P_{C} = (R_{C}, L_{C})$.  The cropping is due to the GPU memory size limitation (16 GB).  In Figure \ref{fig:raster_grid} it is possible to see a graphic description of these pre-processing steps.

The number of initial 256$\times$256 pixel  data-element clips (RGB images)  with their matching binary mask are presented in Table~\ref{tab:data_base}. 

\begin{figure}
    \centering
    \includegraphics[width=0.7\textwidth]{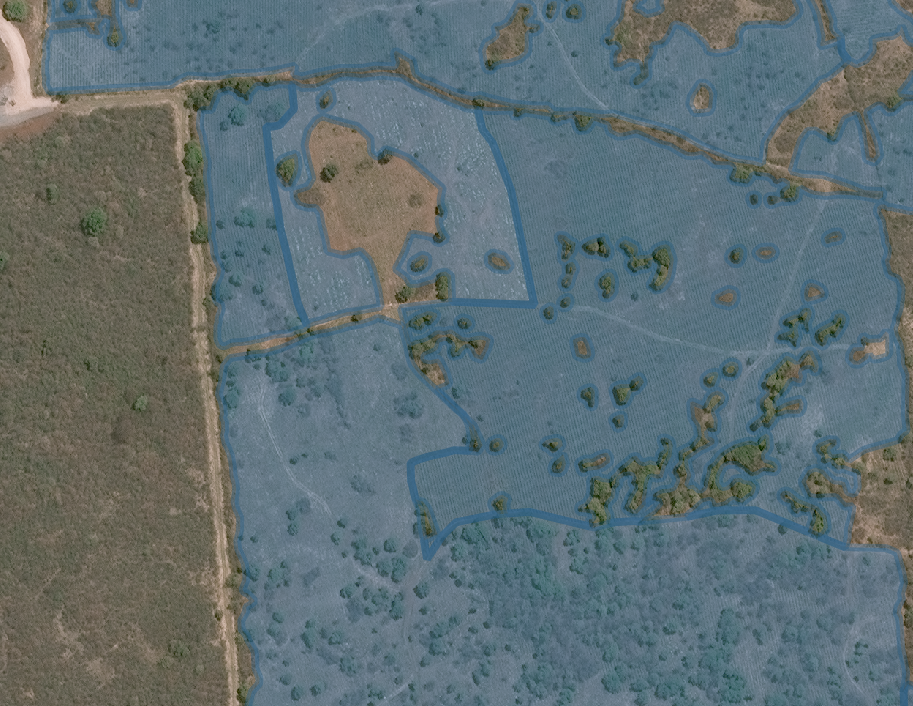}
    \caption{Example of initial phase  agave crop segmentation labels. Note that  some trees, walking  paths, and bushes are  wrongly included in the agave crops labels. }
    \label{fig:init_polygon}
\end{figure}

\begin{table}[h]
    \centering
    \begin{tabular}{lcc}
    \toprule
    Split  data          & Number of elements & Grid ID    \\
    \midrule
    Training      & 127 &  32, 48 and 71  \\
    Validation & 48  &  32, 48 and 71 \\
    Testing       & 208 &  7, 9, 34, 45, 47, 57, 58 and 60 \\
    \midrule
    Total      & 383  \\
    \bottomrule
    \end{tabular}
    \caption{Number of 256 $\times$ 256 images clips and  labels in the initial phase for training, validation, and testing.}
    \label{tab:data_base}
\end{table}
In Figure \ref{fig:init_polygon}, we present an example of the original data (phase 1 data) and  matching labels. It is important to {note that those polygons} (labels) include  not only the agave crops but may also  include trees, water, houses, bushes, etc. The noisy label is one of the main characteristics of  real-world problems (often called \textit{in the wild} \cite{stadelmann2018deepinthewild}).

\subsection{Agave maturity classification}

\begin{figure}[ht]
    \centering
    \includegraphics[width=.8\textwidth]{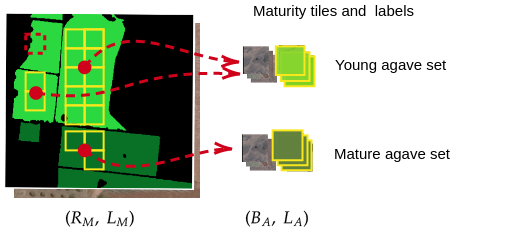}
    \caption{Agave maturity tile extraction. Light green means young agave (up to 2 years) and dark green is mature agave. Yellow bounding-boxes (tiles) represent the locations where a tile was available for extraction, while red bounding-boxes denote unfit tiles.}
    \label{fig:agave_maturity_patches}
\end{figure}

To obtain agave maturity data, we split every tile into $32\times32$ pixels. We defined two classes: young agave (less than or equal to two years of age) and  mature agave (more than two years of age).  Agave maturity classification labels  were prepared by expert photo-interpreters using QGIS and images from different dates were compared to validate evolution of the agave's maturity. Figure \ref{fig:agave_maturity_patches} illustrates the process for extracting the agave maturity tiles (32$\times$32 pixels) and labels $(B_{A}, L_{A}),$ respectively. Note that we selected only those full tiles (yellow bounding-boxes) which are inside the plot limits and that the partial tiles denoted by dashed red lines are not candidates for maturity classification. In this image, light green represents young agave (less than or equal to 2 years) and dark green r mature agave.

Table \ref{tab:agave_maturity_split} presents the number of (32$\times$32 pixel) samples for training, validation, and testing. Note that both classes are  balanced. 
\begin{table}[ht]
\centering
\begin{tabular}{llll}
Maturity set         & Mature & Young & Total \\
\toprule
Training      & 3,904 & 3,904 & 7,808 \\
Validation & 1,735 & 1,735 & 3,470 \\
Testing       & 1,301 & 1,301 & 2,602 \\
\midrule
Total      & & & 13,880 \\
\bottomrule
\end{tabular}
\caption{Number of 32$\times$32 tiles generated for the agave maturity classification training process.}
\label{tab:agave_maturity_split}
\end{table}

\section{Methods}\label{sec:exp}
In this section, we describe the  methods, metrics and tools used to achieve agave crop segmentation and maturity classification. The main method for improving  results was inspired by the AI-Life cycle \cite{desilva-2021-ai-lifecycle} in combination with the active learning process and synthetic satellite image generation.

\subsection{Agave crop segmentation}
\begin{figure}
    \centering
    \includegraphics[width=.8\textwidth]{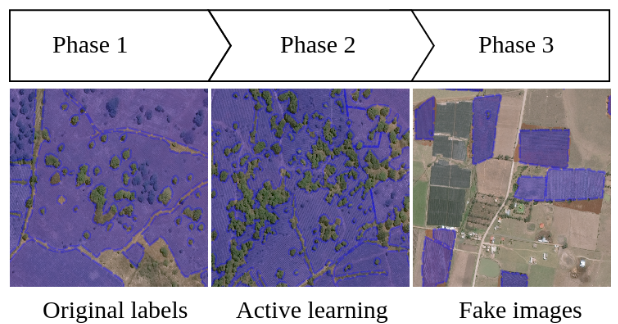}
    \caption{In each phase we present an image to illustrate the original labels (phase 1), improved  labels (phase 2), and fake/synthetic images (phase 3). }
    \label{fig:phases}
\end{figure}

Some  recurrent challenges of deep learning-real world applications are small  training datasets and low-quality labels. As a result, we decided to apply several techniques which could be grouped into three main phases. Figure \ref{fig:phases} illustrates  some examples of images and labels in each phase. In Phase 1, we show a sample of the original labels, where trees, bushes, and footpaths result in additional label noise. In Phase 2, it can be seen that  the improved labels omit trees, bushes, road, footpaths, etc. after using the active learning cycle (human experts+models). The Phase 3 image shows an example of  fake/synthetic agave crops. In the following subsections we explain in more detail each phase of crop segmentation.

\subsubsection{Phase 1: Training, evaluation and testing using original labels}

As noted above,  we trained with two well-known regularization techniques: i) transfer learning, i.e the backbone model has been pre-trained with the COCO dataset, and ii) data augmentation.  After that, we evaluated the model's performance in the test set using DCL and IoU. These results are considered the performance baseline for comparison with our models and data.

\subsubsection{Phase 2: Active learning}

In the second phase, we focused on  active learning. Specifically, experts improved  agave crop  delimitation avoiding inside the crops roads, trees, bushes,  houses, water, etc. This technique also helped to increase the number of  training samples/labels with a DL-model $+$ human workflow. To do this, the inferences from the DL model propose new labeled areas and human-experts correct and approve the new samples with the corresponding labels.

\subsubsection{Phase 3: Synthetic/fake images}

\begin{figure}[h]
    \centering
    \includegraphics[width=0.3\textwidth]{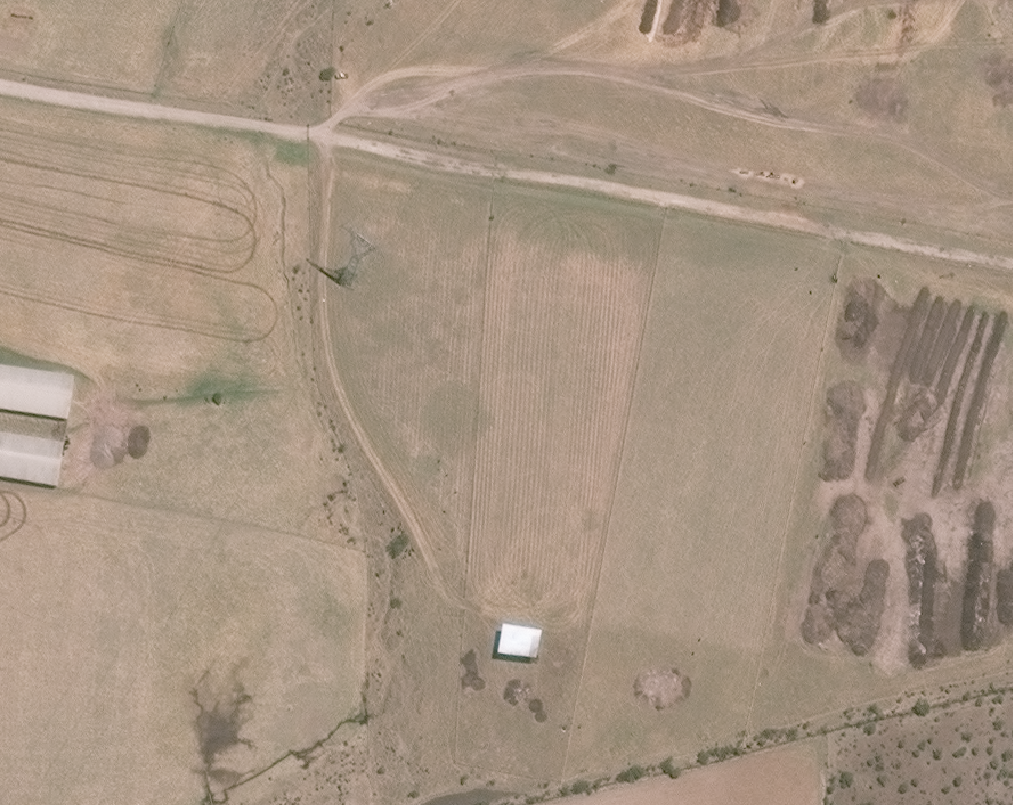}
    \includegraphics[width=0.3\textwidth]{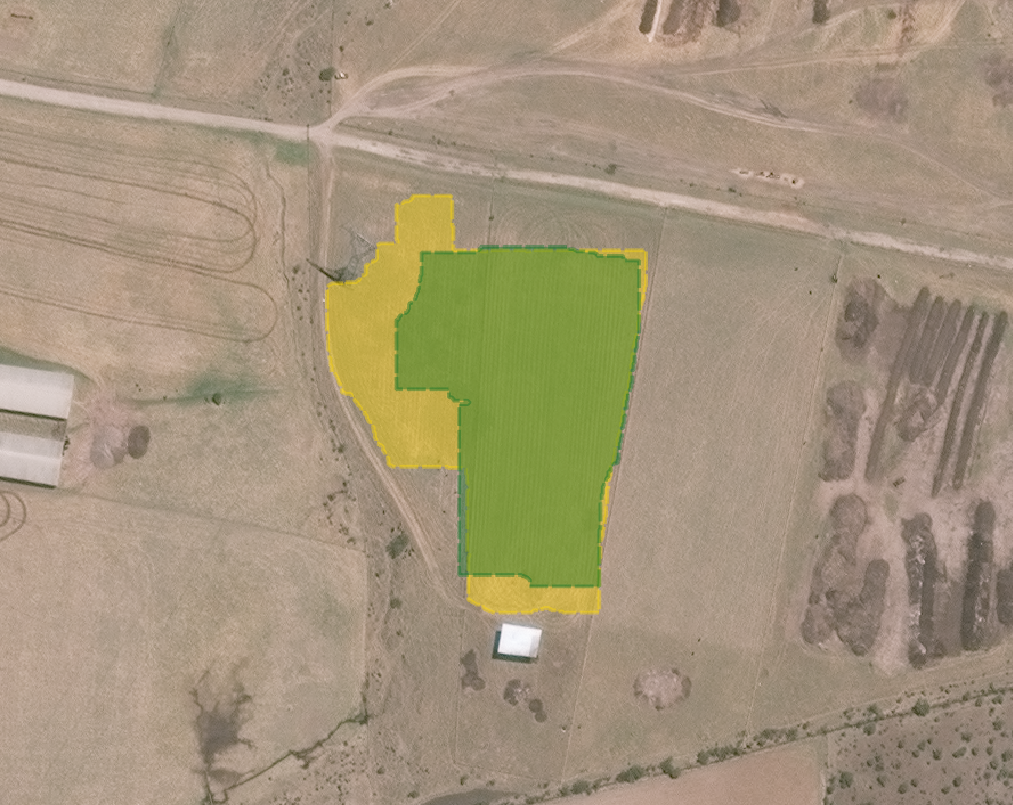}
      \includegraphics[width=0.3\textwidth]{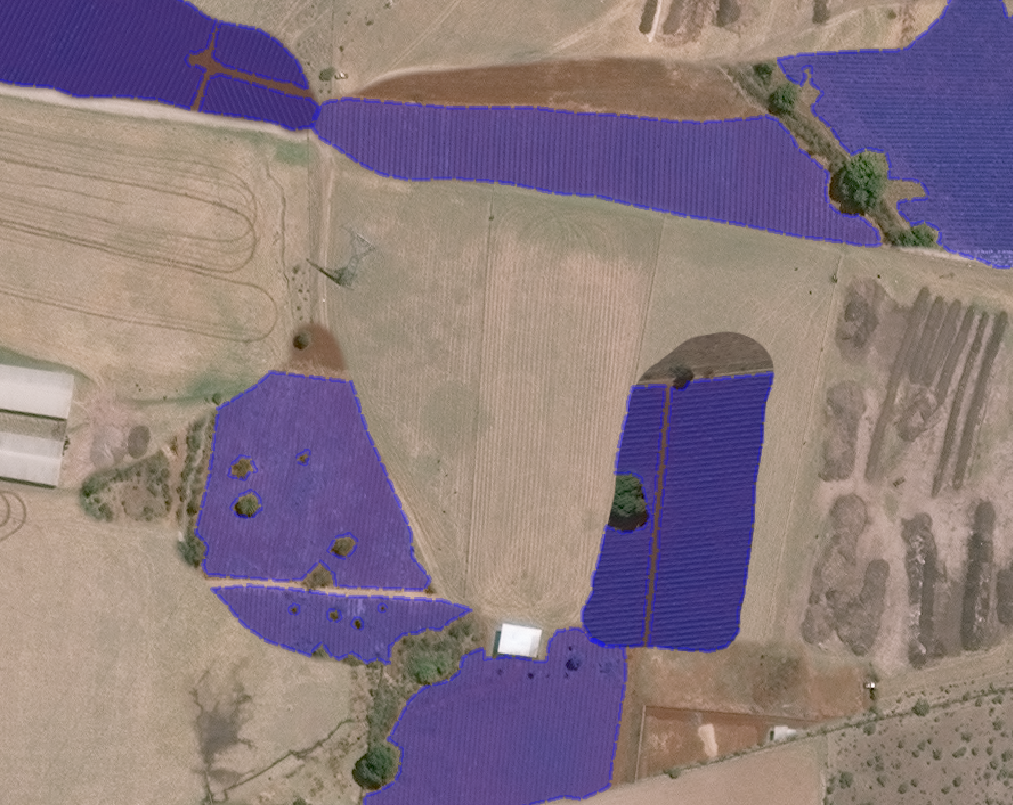}
    \caption{Left: Original image. Center model's inference and label. Right  examples of synthetic/fake images.}
    \label{fig:similar_to_agave}
\end{figure}

In phase 3, we generated synthetic images  based on  the model's most frequent  validation error (out-of-distribution-data) using an GNU Image Manipulation Program \href{https://www.gimp.org/}{GIMP}.  Two examples of the phase 1-2 errors are  presented in the Figure \ref{fig:similar_to_agave}. These examples help  to define the textures  and regions in the synthetic images. 

\subsubsection{Segmentation metrics}
The  metrics selected  to evaluate agave crop segmentation performance models are: Dice Coefficient Loss (DCL) and Intersection Over Union (IoU). The main reason  for using these metrics is due to those metrics being widely used in many semantic segmentation tasks \cite{zhang2020transferability,zhao-dice-loss-2020,maxwell2021accuracy}. Moreover,  IoU  is frequently employed as the cutoff point for deciding whether objects should be classified as True Positives (TP) or False Negatives (FN), with 0.5 as the threshold. In addition, DCL and IoU allow the evaluation of highly unbalance classes~\cite{zhao-dice-loss-2020,zhu-2021-iou}.

DCL is implemented using the  \href{https://smp.readthedocs.io/en/latest/}{Segmentation Models Library}. In this library, DCL is computed based on the Dice Similarity Index  ($DSI$)  defined by  equation \ref{eq:dcl} as follows:
\begin{equation}
     DSI =  \frac{2 |X \cap Y|}{|X| + |Y|} \label{eq:dcl}.
\end{equation}
For Jaccard index or $IoU$  we also used the implementation from \href{https://smp.readthedocs.io/en/latest/}{Segmentation Models Library} which is fully compatible with our Pytorch implementation. $IoU$ Equation \ref{eq:IoU} is defined as follows: 
\begin{equation}
IoU = {{|X \cap Y|}\over{|X \cup Y|}} = {{|X \cap Y|}\over{|X| + |Y| - |X \cap Y|}}. \label{eq:IoU}    
\end{equation}
In  equations \ref{eq:dcl} and \ref{eq:IoU} the two sets $X, Y \in [0,1]$ represent the ground truth label and the model inference value respectively. In addition, $X \cap Y$ represent the pixel-to-pixel intersection, $X \cup Y$ is the union of the two sets, and   $|X|$, $|Y|$  is the  cardinality of the two sets $X$, $Y$, respectively,  i.e, the number of pixels in each set \cite{zijdenbos-dice-1994}. 

\subsubsection{ConvNets models}

The ConvNets used in this work for crop segmentation are Mask R-CNN \cite{maskrcnn}, Unet++ \cite{unetplusplus}, DeepLabV3+ \cite{deeplabv3plus} y FPN \cite{fpn}. In all phases we used transfer learning using COCO \cite{coco} weights, and the same data augmentation: $90^o$ random rotations, horizontal  and, vertical splits in the training set.  The hyper-parameters of the training process are a learning rate of 0.0001, and 50 epochs.  We used an IBM AC-922 server with four V100 GPU (16GB) for all training.

\subsection{Agave maturity classification model}

Agave maturity tile classification was carried out using  pretrained  MobileNetV2 \cite{mobilenet}, ResNet18 \cite{resnet} and EfficientNet B0 \cite{efficientnet}, with a learning rate of 0.0001, an Adam \cite{adam} optimizer, and categorical cross entropy loss. In addition, we used  data augmentation techniques such as horizontal and vertical flips and random rotations up to $90^o$. For all training, we used an IBM AC-922 server with V100 GPU (16GB). To evaluate the model's performance, we used the following:
\begin{eqnarray}
     accuracy &=& \frac{TP+TN}{TP+TN+FP+FN}, \label{eq:acc} \\
      sensitivity &=& \frac{TP}{TP+FN}, \\
      specificity &=& \frac{TP}{TN + FP}, 
      \end{eqnarray}
where $TP =$ True Positive, $TN =$ True Negative, $FP =$ False Negative, and $FP = $ False Positive.

\begin{figure}[h]
    \centering
    \includegraphics[width=.8\textwidth]{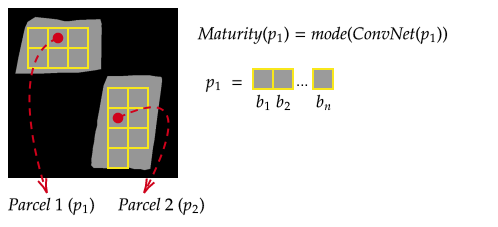}
    \caption{Example of the agave maturity classification computation process. First, we appied  the inference using the ConvNet model in each parcel tile, and then we computed the mode.}
    \label{fig:agave_maturity_inference}
\end{figure}

Figure \ref{fig:agave_maturity_inference} introduces the process for evaluating the maturity of agave parcels. Each parcel tile (yellow box) is classified using the pretrained ConvNet and then computing the mode to assign the maturity of the parcel.

\section{Results and analysis}\label{sec:results}

\subsection{Agave crop segmentation}

Table \ref{tab:data_split} presents the number of elements in train-set, validation-set and test-set in each phase. It is important to highlight   that the number of training and validation samples increase in each phase. Moreover,  our experience hasdemostrated empirically that human/model collaboration reduces the labeling time of human experts. Informally,  we note that the labeling time was reduced  by a range of factor between  3 to 5. 

\begin{figure}[ht]
    \centering
    \includegraphics[width=1.\textwidth]{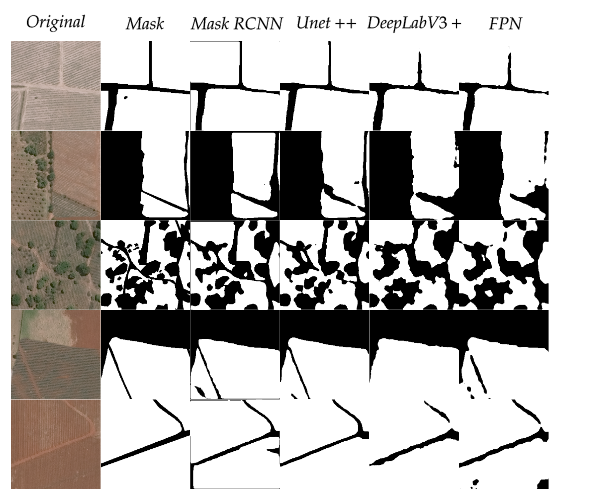}
    \caption{Original RGB images, ground truth labels,  and segmentation results from different models: Mask R-CNN, Unet++, DeepLabV3+  and FPN.}
    \label{fig:matrix_results}
\end{figure}

Figure \ref{fig:matrix_results} shows five test-set examples using the  phase 3 data. The first column presents the {Original} RGB image and the second is ground truth label. The following columns present the segmentation results from the Mask R-CNN, Unet++, DeepLabV3+, and FPN,  respectively. It can be seen  that all  models could detect the agave crops in sample images. However,  due to post-processing   age estimation,   we  consider better the crop split capacity of the Mask R-CNN.

\begin{table}[ht]
    \centering
    \begin{tabular}{lccc}
    \toprule
      Segmentation Set        & Phase 1   & Phase 2   & Phase 3 \\
    \midrule
    Training      & 127 & 182 & 528 \\
    Validation & 48  & 74  & 222 \\
    Testing       & 208 & 208 & 208 \\
    \midrule
    Total      & 383 & 464 & 958 \\
    \bottomrule
    \end{tabular}
    \caption{Number of ($256 \times 256$) clips in training-set, validation-set and test-set in each phase.}
    \label{tab:data_split}
\end{table}

Table \ref{tab:cnn_results}  presents each model's agave crop segmentation  performance numerically. The mean   $IoU$ values increase from  initial phase  from   0.725  to 0.905  in the final phase (higher values is better) and $DCL$  mean value from the initial phase 0.175 to 0.047  in the final phase (lower values are better). Note that the significantly increase in the segmentation performance is from  phase 2 to  phase 3 where synthetic images and the active learning process are  included only in the training set. Comparing the four ConvNet models the MaskRCNN models achieves the best segmentation performance against  other ConvNets models.  As a result, we consider the Mask R-CNN to have the best qualitative and quantitative results.


\begin{table}[ht]
\centering
\begin{tabular}{lccccccc}
\toprule
\multirow{2}{*}{Architecture} & \multicolumn{2}{c}{{Phase 1}} & \multicolumn{2}{c}{{Phase 2}} & \multicolumn{2}{c}{{Phase 3}} & \multirow{2}{*}{Param(M)} \\ 
           & $IoU$  & $DCL$ & $IoU$  & $DCL$ & $IoU$  & $DCL$ &    \\
\midrule
Mask RCNN \cite{maskrcnn}  & \textbf{0.76} & \textbf{0.15}      & 0.74 & 0.18      & \textbf{0.91} & \textbf{0.047}     & 49.9 \\
Unet++ \cite{unetplusplus}     & 0.74 & 0.18      & \textbf{0.76} & \textbf{0.17}      & 0.91 & 0.05      & 48.5 \\
DeepLabV3+ \cite{deeplabv3plus} & 0.70 & 0.19      & 0.72 & 0.18      & 0.90 & 0.05      & 26.1 \\
FPN \cite{fpn}       & 0.70 & 0.18      & 0.71 & 0.17      & 0.90 & 0.054      & 25.6 \\
\midrule
Mean     & 0.725 & 0.175      & 0.733 & 0.175      & 0.905 & 0.050      & -- \\
\bottomrule
\end{tabular}
\caption{Comparison between the four models  and the three  data/label improvement phases.}
\label{tab:cnn_results}
\end{table}

The most important observation from Tables  \ref{tab:cnn_results}, and  \ref{tab:data_split}  is that the active learning and synthetic images not only increase the  number of trainable images through an effective human-model collaboration but also increase the  data-label quality  and models performance.

\subsection{Agave maturity classification}
\begin{figure}[h
]
    \centering
    \includegraphics[width=\textwidth]{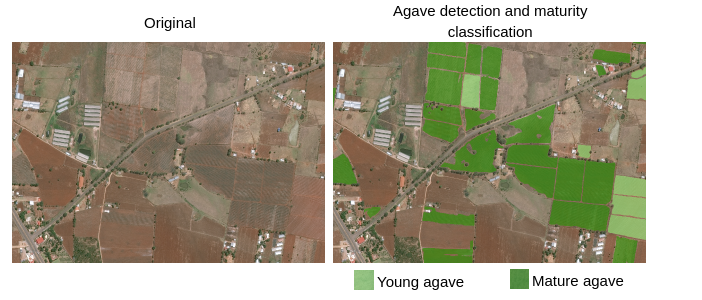}
    \caption{Agave detection and maturity classification example.}
    \label{fig:r_agave_age}
\end{figure}

The quantitative results of agave maturity classification  performance in the test set are presented in Table \ref{tab:age_class_report}. In our opinion the best trade-off between performance and number of trainable parameters is MobileNetV2.  An example  of the qualitative results  is presented in Figure \ref{fig:r_agave_age}, with the original image and the classification areas in light green (young)  and  dark green (mature). We wish to underscore that this model helped us realize that the segmentation model made more mistakes in the  parcels of   young agave due to the similarities between cultivation line patterns and soil color.

\begin{table}[h]
    \centering
    \begin{tabular}{llllll}
    \toprule
    Model          & Accuracy &  Sensitivity & Specificity & Loss  & Params (M) \\
    \midrule
    MobileNetV2 \cite{mobilenet}    & 0.97   & 0.98  & 0.96   & 0.079 & 2.2        \\
    Resnet18 \cite{resnet}      & 0.97   & 0.97  & 0.97   & 0.084 & 11.2       \\
    EfficientnetB0 \cite{efficientnet} & 0.97   & 0.99  & 0.95   & 0.101 & 4          \\
    \bottomrule
    \end{tabular}
    \caption{Agave maturity classification report on three ConvNet architectures.}
    \label{tab:age_class_report}
\end{table}

\section{Conclusions}\label{sec:conclusions}

Our work is the first to implement deep learning techniques for these  agave tasks on this scale. Moreover, this  work  presents a novel strategy for large-scale DL  agave-parcel segmentation and maturity classification  using very high-resolution satellite imagery.  The proposed strategy  solves typical DL real-word  problems such as a  lack of data, low-quality labels, poor model performance and highly imbalanced data in the very specific context of agave. This strategy combines  active learning,  and data-centric (synthetic image/label generation) techniques. According to the results, this strategy allows    significantly increase  the number of samples/labels and segmentation performance in less time and with effort on the part of  photointerpreters. This could have a major impact on monitoring and mapping tasks, reducing the time required to digitize agave crops compared to traditional human only methods. Regarding the DL models, Mask-R CNN  had the best trade-off between IoU, DL and the number of trainable parameters.  This agave  maturity classification method is  95 \%, accurate; however, we detected more error (biases) in the young agave class.   These models can be used to monitor large regions, on the order of thousands of squared kilometers, and can contribute to reducing  social and environmental problems resulting from the agave-tequila production chain. As a future work, we are planning to use other satellite imagery to detect  agave crops. 
 
\textbf{Acknowledgments.} The authors would like to thank CONACYT, and SEMADET for their valuable help.

\textbf{Data availability.} The data that support the findings of this study are available from Maxar but restrictions apply to the availability of these data, which were used under license for the current study, and so are not publicly available. Data are however available from the authors upon reasonable request and with permission of Maxar. The corresponding labels, models and code are available from the corresponding author upon reasonable request.

\section*{Declarations}
\textbf{Conflict of interest.} The authors declare that they have no conflict of interest.

\bibliographystyle{ieeetr}

\bibliographystyle{ieeetr}







\end{document}